%% file: IEEE-conference-template-062824.tex
\documentclass[conference]{IEEEtran}
\IEEEoverridecommandlockouts

\usepackage{cite}
\usepackage{amsmath,amssymb,amsfonts}
\usepackage{algorithmic}
\usepackage{graphicx}
\usepackage{textcomp}
\usepackage{xcolor}
\usepackage{algorithm}
\usepackage{xcolor}
\usepackage{caption}
\usepackage{booktabs}
\def\BibTeX{{\rm B\kern-.05em{\sc i\kern-.025em b}\kern-.08em
    T\kern-.1667em\lower.7ex\hbox{E}\kern-.125emX}}
\begin{document}

\title{A Faster and More Reliable
Middleware for Autonomous Driving Systems}

\author{\IEEEauthorblockN{1\textsuperscript{st} Yuankai He}
\textit{University of Delaware}
\and
\IEEEauthorblockN{2\textsuperscript{rd} Weisong Shi}
\textit{University of Delaware}}

\maketitle

\input{sections/0_Abstract}
\input{sections/1_Introduction}
\input{sections/2_BackgroundInfo}

\input{sections/3_SystemDesign}

\input{sections/4_Experiment}
\input{sections/5_Conclusion}

\bibliographystyle{IEEEtran}
\bibliography{references}
\end{document}

%% file: sections/0_Abstract.tex
\begin{abstract}
Ensuring safety in high-speed autonomous vehicles depends on rapid control loops and tightly bounded delays from perception to actuation. Many open-source autonomy systems depend on ROS~2 middleware, but when multiple sensor and control nodes share a single computing unit, ROS~2 and its DDS transports introduce significant (de)serialization, data copying, and discovery overheads, eroding the available time budget. We present \emph{Sensor-in-Memory (SIM)}, a shared-memory transport specifically designed for intra-host pipelines in autonomous vehicles. SIM keeps sensor data in their native memory layouts (e.g., \texttt{cv::Mat}, PCL), uses lock-free bounded double buffers that overwrite old data to prioritize freshness, and requires only four lines of code to integrate into ROS~2 nodes. Unlike traditional middleware, SIM operates beside ROS~2 and is optimized for applications where data freshness and minimal latency outweigh guaranteed completeness. SIM incorporates sequence numbers, a writer heartbeat, and optional checksums to provide data ordering, liveness, and basic integrity. On NVIDIA Jetson Orin Nano, SIM reduces data transport latency by up to 98\% compared to ROS~2 zero-copy DDS such as FastRTPS and Zenoh, lowers mean latency by about 95\%, and sharply narrows p95/p99 outlier latency by around 96\%. In tests on a production-ready L4 vehicle running Autoware.Universe, SIM increased localization frequency from 7.5\,Hz to 9.5\,Hz. When applied across all latency-critical modules, SIM cuts average perception-to-decision latency from 521.91\,ms to 290.26\,ms, resulting in a 13.6\,ft (4.14\,m) shorter emergency braking distance at 40\,miles/hr (64\,km/hr) on dry concrete.
\end{abstract}

%% file: sections/1_Introduction.tex
\section{Introduction}\label{sec:intro}

Autonomous vehicles (AVs) are safety-critical. As speed increases, the \emph{allowable perception-to-decision latency shrinks} because the vehicle travels farther each millisecond. At 40\,mph ($\sim$17.9\,m/s), every 100\,ms of delay adds $\sim$1.79\,m to reaction distance. On a vehicle running the open-source Autoware.Universe stack, we measure a mean perception-to-decision time of \textbf{521.91\,ms}, with \textbf{369.45\,ms} in ROS~2 publish/subscribe paths---showing that the communication substrate can dominate end-to-end delay and tail behavior~\cite{liu20214c}.

Open-source autonomy stacks, such as Autoware.Universe~\cite{RefWorks:kato2018autoware} and earlier Apollo releases~\cite{fan2018baidu}, commonly use ROS~2~\cite{doi:10.1126/scirobotics.abm6074} with DDS for intra-vehicle messaging. Publish--subscribe improves modularity and reuse, but \emph{(de)serialization}, \emph{redundant copies}, and \emph{dynamic discovery} add latency and jitter, especially with high-rate cameras and LiDAR. On embedded platforms with limited CPU and memory, these DDS-induced costs cap control-loop frequency and compress safety margins.

We introduce \emph{Sensor-in-Memory (SIM)}, a domain-specific \emph{shared-memory transport} and API for \emph{intra-host} (single-ECU) AV pipelines. SIM keeps sensor data in \emph{application-native layouts} (e.g., \texttt{cv::Mat}, PCL), removes (de)serialization and format conversion, and uses \emph{bounded double buffers} with \emph{overwrite-on-update} to favor \emph{freshness over completeness}. A \emph{lock-free} reader/writer interface yields predictable handoffs. Lightweight safety hooks---\emph{sequence identifiers} (ordering), \emph{writer heartbeats} (liveness), and an \emph{optional checksum} (basic integrity)---reduce the risk of stale or corrupted reads. SIM integrates into existing ROS~2 nodes with $\approx$4 lines per pub-sub pair, preserves open-source tooling, and coexists with DDS/Ethernet for inter-ECU links. 

\textbf{Scope:} intra-host pipelines where low latency and freshness are the primary objectives.

We evaluate SIM on NVIDIA Jetson Orin Nano and on a full-scale production-ready L4 vehicle against widely used \emph{open-source, ROS~2-compatible zero-copy transports}: Fast DDS in shared-memory configuration~\cite{RefWorks:eprosimafastrtps}, and Zenoh~\cite{RefWorks:corsaro2023zenoh:}. Across camera and LiDAR pipelines, SIM lowers end-to-end latency---\textbf{up to 98\% lower maximum}, \textbf{$\approx$95\% lower mean}---and \textbf{reduces p95/p99 tails by $\approx$96\%}. On Autoware.Universe, these gains raise application throughput (\emph{NDT matching} 7.5\,Hz $\rightarrow$ 9.5\,Hz) and cut average perception-to-decision latency (521.91\,ms $\rightarrow$ 290.26\,ms). The shorter reaction time implies a \textbf{13.6\,ft (4.14\,m)} reduction in emergency-braking distance at \textbf{40\,mph (64\,km/h)} on dry concrete pavement.

\noindent\textbf{Contributions}
\begin{itemize}
  \item \textbf{Safety-oriented bottleneck analysis} of intra-vehicle ROS~2/DDS paths in the Autoware autonomy stacks under high-rate sensing.
  \item \textbf{Design and implementation} of \emph{SIM}, a native-layout, overwrite-first, bounded shared-memory transport with safety-aware lifecycle primitives (ordering, liveness, integrity).
  \item \textbf{Low-friction ROS~2 integration} ($\approx$4 lines per pub-sub pair) that preserves open-source ecosystem compatibility.
  \item \textbf{Empirical comparison} on embedded and vehicle platforms against \emph{Fast DDS (zero-copy SHM)}and \emph{Zenoh}~\cite{RefWorks:eprosimafastrtps, RefWorks:corsaro2023zenoh:}, showing large mean/max reductions and \textbf{$\approx$96\%} p95/p99 tail reduction, plus application-level gains in Autoware.Universe.
\end{itemize}

Section~\ref{sec:background} reviews ROS-based AV dataflow and details DDS overhead and jitter~\cite{liu20214c}. Section~\ref{sec:design} presents SIM’s design and API. Section~\ref{sec:experiment} reports camera/LiDAR results versus Fast DDS (zero-copy) and Zenoh on embedded and vehicle platforms. Section~\ref{sec:conclusion} concludes the paper.

%% file: sections/2_BackgroundInfo.tex
\section{Data Flow and the Role of DDS in an Autonomous Vehicle}\label{sec:background}

\begin{figure*}[!htbp]
    \centering
    \includegraphics[width=2.0\columnwidth]{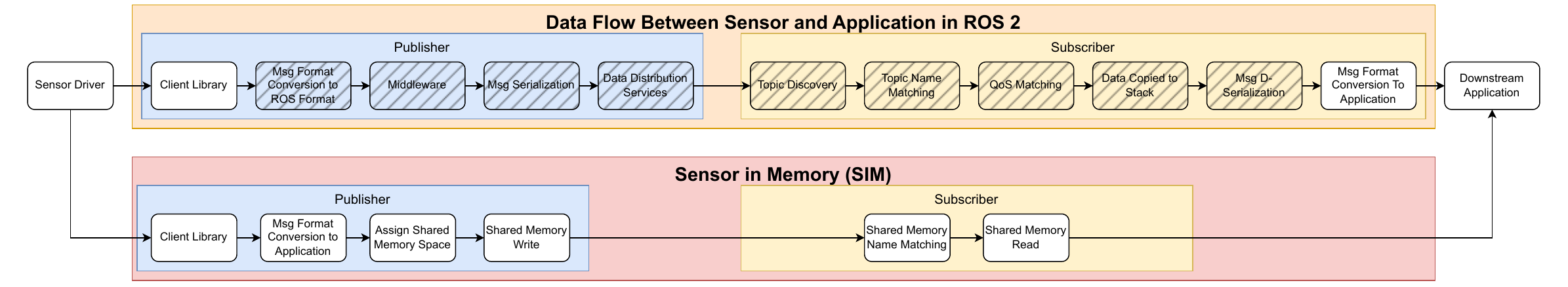}
    \caption{ Top: Data Flow and the Role of DDS in an Autonomous Vehicle \\ Bottom: SIM Overview}
    \label{fig:avoverview}
\end{figure*}

\subsection{ROS~2/DDS Intra-Host Data Path}
In open-source AV stacks, ROS~2 with DDS provides publish--subscribe messaging and QoS control for perception, localization, planning, and control~\cite{doi:10.1126/scirobotics.abm6074}. As sketched in Fig.~\ref{fig:avoverview} (top), sensor outputs are converted to ROS message types, serialized by DDS, transported, deserialized, and rematerialized into algorithm-native structures. Each step---type conversion, (de)serialization, queueing, and buffer handoff---adds CPU work and delay; with high-rate cameras/LiDAR and embedded SoCs, the accumulated cost eats into the perception-to-decision budget and stresses latency predictability.

\subsection{What Measurements Show}
Empirical studies already show that message size, executor choice, and QoS settings leads to increases in mean and tail latency in ROS~2/DDS pipelines~\cite{RefWorks:2016exploring,RefWorks:jiang2020message,RefWorks:2021performance,RefWorks:2021latency,RefWorks:2022end-to-end,RefWorks:kouril2024performance}. A recurring observation is that (de)serialization and redundant copying dominate at high throughput, and that p95/p99 behavior is as consequential as the mean for safety-critical loops.

\subsection{Tuning Within ROS~2/DDS (and Its Limits)}
Response-time analysis and runtime tuning, such as profiling, executor variants, priority/affinity scheduling, and parameter auto-tuning, can reduce jitter~\cite{RefWorks:arafat2022response,RefWorks:2024dynamic,RefWorks:2023impact}. However, these methods retain DDS abstractions; conversions and copies on the sensor-to-application largely remain, and tail latency often stays sensitive to background load~\cite{RefWorks:2022end-to-end,RefWorks:2021latency}.

\subsection{Zero-Copy and Shared-Memory Communication}
A complementary line of work shows that bypassing copies lowers latency and CPU load. Wu et al.\ and Bell et al.\ quantify the benefits of shared memory and zero-copy paths in latency-sensitive deployments~\cite{RefWorks:2021oops!,RefWorks:2023hardware}. Systems such as \emph{TZC} and \emph{ROS-SF} further reduce copying and OS crossings~\cite{RefWorks:wang2019tzc:,RefWorks:wang2022ros-sf:}. Several ROS~2 zero-copy paths (e.g., Fast DDS data-sharing with \emph{loaned messages}) reduce copies \emph{within the middleware boundary}, but they apply primarily to \emph{plain/fixed-size} types and require adopting loan/return APIs; applications commonly convert ROS messages to algorithm-native layouts (e.g., OpenCV/PCL) before processing, which reintroduces conversion cost. DDS/RTPS exposes ordering and liveliness via QoS, yet per-frame \emph{application-level} guards (explicit sequence tags, writer heartbeats, checksums) are typically left to the application~\cite{doi:10.1126/scirobotics.abm6074}. These gaps motivate an AV-specific, lifecycle-aware, native-layout design.

\subsection{Alternatives Beyond ROS~2}
Apollo \emph{CyberRT} provides custom scheduling and IPC for high-throughput pipelines, and \emph{AUTOSAR Adaptive} offers a service-oriented architecture with certification pathways~\cite{ap,RefWorks:fürst2016autosar}. These target production ecosystems and are proprietary or tightly bound to specific stacks, limiting generalizability and independent evaluation in open-source contexts. For comparability and reproducibility, we center our study on ROS~2--based pipelines.

\subsection{Requirements for an AV Intra-Host Transport}
Evidence above implies concrete requirements:
\paragraph{R1: Native layouts, no (de)serialization on the sensor-to-application path.} Remove conversion costs by publishing/consuming algorithm-native representations~\cite{RefWorks:2021oops!,RefWorks:2023hardware}.  
\paragraph{R2: Freshness-first, bounded sharing.} Deterministically drop stale frames to keep reaction distance small under load.  
\paragraph{R3: Predictable handoff with minimal CPU work.} O(1) copies and a lock-free common case~\cite{RefWorks:2022end-to-end,RefWorks:2021latency}.  
\paragraph{R4: Safety-aware lifecycle.} Ordering, liveness, and basic integrity for multi-sensor, multi-threaded pipelines~\cite{RefWorks:2024rtex:,RefWorks:2021picas:}.  
\paragraph{R5: ROS~2 compatibility.} Minimal code changes and coexistence with DDS for inter-ECU links~\cite{doi:10.1126/scirobotics.abm6074}.

\subsection{Positioning of SIM}
As shown in Fig.~\ref{fig:avoverview} (bottom), SIM places a shared-memory buffer on the sensor-application data path, bypassing DDS while preserving ROS~2 integration. Data producers write application-native layouts; bounded double-buffers with overwrite-on-update enforce freshness (R2) and O(1) handoffs (R3); and sequence identifiers, writer heartbeats, and an optional checksum provide ordering, liveness, and basic integrity (R4). Integration occurs within existing ROS~2 nodes with $\approx$4 lines of code, leaving inter-ECU DDS paths unchanged (R5). The goal is not only lower averages but tighter p95/p99 latency in the perception-to-decision loop, which is directly tied to lost stopping margin.

%% file: sections/3_SystemDesign.tex
\section{System Design} \label{sec:design}
\begin{figure*}[htbp]
    \centering
    \includegraphics[width=2.0\columnwidth]{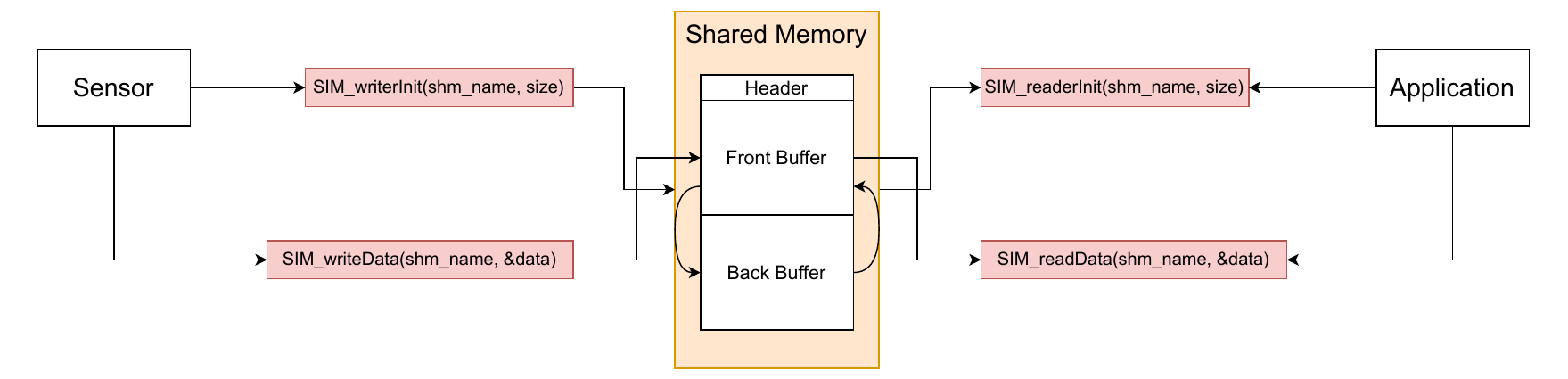}
    \caption{SIM maps shared memory for each sensor, using unique identifiers, such as /camera\_front, with dynamically sized buffers based on sensor specifications. Each sensor has dedicated regions without disrupting others. Library APIs like ${\tt SIM\_writerInit(name, size)}$, ${\tt SIM\_writeData(data, size)}$, ${\tt SIM\_readerInit(name, size)}$, ${\tt SIM\_readData(data, size)}$, and ${\tt SIM\_destroy(name)}$ are created facilitate memory management and data flow. The writer only flips the \texttt{front\_idx} when it has finished writing to ensure no data is torn.}
    \label{fig:simarch}
\end{figure*}

\subsection{Role and Scope}
SIM shortens the \emph{sensor-to-application} path in open-source AV stacks by replacing the ROS~2/DDS message/(de)serialization/copy chain with a preallocated shared-memory data plane. The design is strictly \emph{intra-host}: it carries high-rate frames between local producers and consumers (e.g., LiDAR$\rightarrow$localization, camera$\rightarrow$obstacle detection), while inter-ECU links remain on ROS~2/DDS or Ethernet. Consistent with \S\ref{sec:background}, SIM preserves native algorithm layouts, enforces freshness-first and bounded sharing, provides constant-time publication and consumption without locks, exposes per-frame ordering and liveness, and remains compatible with ROS~2 tooling and deployments.

\subsection{Architecture}
As shown in Fig \ref{fig:simarch}, each real-time stream owns a \emph{named shared-memory region} that is created once and reused across processes. Regions are sized at initialization from sensor configuration so they can hold the \emph{maximum} frame the sensor can produce. For cameras, resolution and format are stable, yielding a fixed byte size per frame. For LiDAR, deployments select a \texttt{MAX\_POINTS} upper bound and allocate capacity accordingly. The region contains a small header and two payload buffers (double buffering). In our LiDAR instance, the header stores \texttt{front\_idx} (the published buffer, as an atomic integer), per-buffer \texttt{frame} numbers and \texttt{timeStamp} values, and a \emph{published length} that records the size of the currently published frame. The payload holds \texttt{PointXYZ data[2][MAX\_POINTS]}. Image streams store width, height, channels, stride, and depth so the payload is directly viewable as a \texttt{cv::Mat}. All memory is preallocated and mapped once with POSIX shared memory; steady-state operations do not allocate or make system calls. Payloads are laid out exactly as algorithms expect---PCL arrays for LiDAR; \texttt{cv::Mat} for images---so there is no message conversion or (de)serialization on the sensor-to-application path.

\subsection{Publish/Consume Protocol}
A single writer publishes by filling the non-published buffer and then flipping \texttt{front\_idx} atomically. Concretely, the writer selects the back buffer, writes the new frame into \texttt{data[back]}, sets the \emph{published length} to the number of points (or bytes) actually written for this frame, updates \texttt{frame[back]} and \texttt{timeStamp[back]}, and finally makes the frame visible with a store that uses release ordering on \texttt{front\_idx}. Readers begin by loading \texttt{front\_idx} with acquire ordering, check whether the \texttt{frame} number differs from the one they last consumed, and if so read the \emph{published length} and copy exactly that many elements from the selected buffer into their workspace before updating their local \texttt{last\_frame}. Because the writer commits only after finishing the back buffer and the length field, and readers always acquire the published index before reading, a reader either copies the new complete frame or the previously published one---never a torn frame and never beyond the written bounds. If producers outpace consumers, intermediate frames are intentionally skipped. That choice favors \emph{freshness over completeness}, which is the safer policy for control at speed.

\subsection{Capacity Provisioning and Variable Effective Size}
SIM provisions memory to the \emph{maximum} the sensor can produce and supports frames that are \emph{smaller than capacity} without padding. The header’s \emph{published length} records the actual number of points (for LiDAR) or bytes/pixels (for images) written for the currently published frame. Writers set this length before publishing the new \texttt{front\_idx} with release semantics; readers observe both the index and the matching length under acquire semantics and therefore never read outside the address range. This policy removes per-frame size races while preserving constant-time publication and consumption.

\subsection{Correctness and Timing Properties}
We assume a single writer per stream, a bounded capacity fixed at initialization, and crash-stop processes. Capacity is respected because the published length is always $\le$ the region’s capacity. Publication order is well defined because only the buffer being published ever has its \texttt{frame} incremented. Visibility is guaranteed by writing the payload, its metadata, and the published length before the published index becomes visible to readers. From these invariants it follows that readers never observe torn frames or read past the end of the written data; readers do not wait for one another or for the writer; and memory usage and queuing are bounded to two buffers. The steady-state handoff time for a frame of size $B$ bytes is well approximated by
$
T_{\text{handoff}} \;=\; \frac{B}{BW_w} \;+\; T_{\text{publish}} \;+\; \frac{B}{BW_r},
$
where $BW_w$ and $BW_r$ are sustained \texttt{memcpy} bandwidths of the writer and reader cores and $T_{\text{publish}}$ comprises a small number of atomic operations. There is no term for discovery, dynamic allocation, or (de)serialization, which explains the lower means and tighter p95/p99 observed in \S\ref{sec:experiment}.

\subsection{Fault Handling and Observability}
Shared regions are kernel-backed and \emph{named}, so a crashed process can reattach by name and resume operation without rebooting the stack. Readers detect producer stalls by comparing the most recent \texttt{timeStamp} against a per-stream deadline. A minimal diagnostics surface reports drops and the maximum inter-publish interval to aid tuning; during bring-up a per-buffer checksum can be enabled to catch corruption and then disabled in steady operation to save cycles.

\subsection{Placement and OS Configuration}
Headers and payloads are cache-line aligned, and writer/reader threads are pinned to the same NUMA node as the region. Readers map the region read-only; writers map read--write. High-rate streams may lock pages in memory to avoid paging.

\subsection{ROS~2 Compatibility and Deployment}
SIM is a small C++ library that preserves ROS~2 workflows. In latency-critical segments, a developer replaces a publisher/subscriber pair with SIM calls; in practice this change is on the order of four lines per node. Bridge processes expose SIM streams to remote consumers (SIM$\rightarrow$DDS) or mirror remote inputs into local shared memory (DDS$\rightarrow$SIM), which allows incremental rollout without disturbing inter-ECU communication.

\subsection{Design Choices and Scope}
We use double buffering rather than rings because queues increase delay and tail variance and optimize for completeness rather than freshness. Readers copy out by default because that keeps them independent and wait-free; heavier modules can add a borrowed-pointer variant without changing the publication protocol. SIM assumes a single writer per stream and remains intra-host; multi-producer and cross-host transport continue to use ROS~2/DDS.

%% file: sections/4_Experiment.tex
\section{Experiment Setup and Results} \label{sec:experiment}


\subsection{Experiment Setup}


\begin{table*}[!ht]
    \centering
    \caption{Experiment Setup}
    \begin{tabular}{@{}r|c|c|c|c|c @{}}
        \hline
        & CPU Architecture & Max CPU Frequency & CPU Core & RAM & Max RAM Frequency \\
        \hline
        Orin Nano & Arm® Cortex A78AE v8.2 64-bit  & 2.2 GHz & 8 & LPDDR5 8GB & 3200 MHz \\
        \hline
        Autonomous Vehicle & x86\_64 Xeon E5-2667 & 3.2 GHz & 32 & DDR4 128GB & 2400 MHz \\
        \hline
    \end{tabular}
    \label{tab: setup}
\end{table*}%

To evaluate SIM’s mean latency and tail improvements, we benchmark on two compute platforms: (1) an NVIDIA Jetson Orin Nano (embedded, small-scale AV research) and (2) a state-of-the-art autonomous vehicle. Platform specifications are in Table~\ref{tab: setup}. We report CPU specs only, as the evaluated ROS~2 paths do not use the GPU.

We compare SIM against three ROS~2 middleware backends—Fast DDS (FastRTPS), and Zenoh—selected for their adoption and complementary capabilities (high-performance DDS, shared-memory zero-copy IPC, and a lightweight edge-to-cloud protocol, respectively). All backends use identical QoS (queue depth\,=\,1, best-effort), fixed CPU affinity, the same governor, and the same scheduling policy.

This dual-platform design captures both a resource-constrained edge device (Orin Nano) and a high-bandwidth, multi-sensor AV.

Each setup evaluates two publisher–subscriber topologies: (i) one writer to one reader and (ii) one writer to ten readers, covering low-contention and fan-out scenarios. We report maximum, mean, minimum latency, p95/p99, and standard deviation. Latency measures transport time from immediately before \texttt{publish()} to message delivery; camera and LiDAR streams run end-to-end from hardware to application. Sensor models and rates differ per platform due to compute constraints; specifications appear in Table~\ref{tab:sensors}. All ROS~2 sensor drivers timestamp just before the publish call (not at first-pixel/first-point capture).


\begin{table}[!t]
    \centering
    \caption{Sensor configurations}
    \resizebox{0.5\textwidth}{!}{
    \begin{tabular}{@{}l|l|c|c@{}}
        \toprule
        Platform & Sensor (type) & Rate & Payload \\
        \midrule
        Orin Nano & Orbbec Astra Pro (camera) & 30\,FPS & $640{\times}480$ \\
                  & Unitree 4D L1 (LiDAR, 18~ch) & 20\,Hz & 2{,}160 pts/frame \\
        \midrule
        AV        & Basler Pylon (camera) & 42–120\,FPS & $1920{\times}1200$ \\
                  & Pandar64 (LiDAR, 64~ch) & 10\,Hz & 115{,}200 pts/frame \\
        \bottomrule
    \end{tabular}
    }
    \label{tab:sensors}
\end{table}

\subsection{Goal and Hypotheses}
We test whether the design in \S\ref{sec:design}---native layouts with no (de)serialization (R1), bounded freshness-first sharing (R2), and constant-time atomic publication/consumption (R3)---reduces transport latency and tail percentiles on embedded hardware and lowers system scheduler load.
\begin{itemize}
  \item \textbf{H1 (means \& tails):} For LiDAR and camera streams, SIM yields lower \emph{mean}, \emph{p95}, and \emph{p99} transport latency than ROS~2 baselines configured for zero-copy.
  \item \textbf{H2 (robustness under load):} Under 1W$\rightarrow$10R, SIM’s mean, p95, and p99 latency remains under the baseline.
\end{itemize}

\subsection{Methodology}
\textbf{What we measure.} \emph{Transport latency only}: elapsed time from immediately after the writer publishes a frame to immediately after the reader receives it; sensor capture and downstream compute are excluded. The writer stamps each frame after publish; the reader subtracts that stamp on delivery. Clock source: \texttt{CLOCK\_REALTIME}; measured timestamp overhead on Orin Nano: ${20}~\mu$s.\footnote{We verified alternative clocks; \texttt{CLOCK\_REALTIME} had the lowest overhead and variance on our Orin build.}

\textbf{Data collection.} For each configuration, we collect \textbf{5} runs $\times$ \textbf{1000} frames (discard the first \textbf{100} frames). We report min, mean, p95, p99, max, and standard deviation. CPUs are pinned; clocks fixed; identical QoS (depth=1, best-effort) and executor settings across stacks; zero-copy/loaned paths enabled where supported.

\subsection{Results on Jetson Orin Nano}

\textbf{Transports and workloads.} We compare \emph{SIM with elevate scheduling priority to 99} and \emph{SIM without scheduling priority} to \emph{Fast DDS (Zero-Copy)} and \emph{Zenoh}. Streams on Orin Nano (Table~\ref{tab: setup}): indoor LiDAR (18 channels @ 20\,Hz; \texttt{PointXYZ[]} with \textbf{2,160} points/frame) and camera (640$\times$480 @ 30\,FPS; \texttt{cv::Mat}). Concurrency: 1W$\rightarrow$1R and 1W$\rightarrow$10R.

\subsubsection{LiDAR Results (18\,ch @ 20\,Hz)}



\paragraph{1W$\rightarrow$1R.}

\begin{figure}[htbp]
    \centering
    \includegraphics[width=1.0\columnwidth]{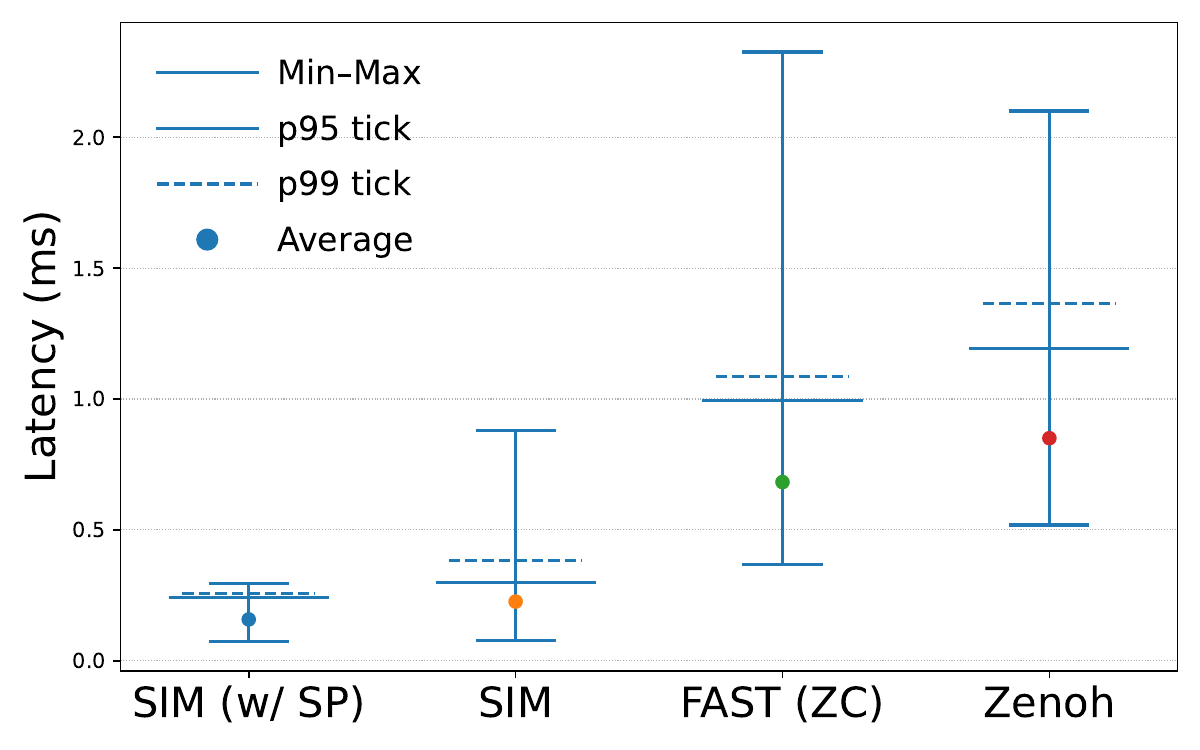}
\caption{LiDAR latency (1W$\rightarrow$1R, linear-scale), Orin Nano. Whiskers show \emph{min–max}; filled dot is the \emph{mean}; long tick is \emph{p95}; short dashed tick is \emph{p99}. Transport-only; best-effort QoS; zero-copy enabled; pinned threads; fixed clocks; No scheduling priority elevation. \textbf{Mean (95\% CI across runs, $n=5$):} SIM 0.226 [0.219, 0.234]; Fast DDS 0.685 [0.650, 0.720]; Zenoh 0.839 [0.773, 0.905]. When scheduling priority is set to 99 with \emph{SCHED\_FIFO}, SIM's mean latency decreases to 0.158 [0.150, 0.164].}
\label{fig:orin_lidar_whisker_minmax1w1r}
\end{figure}

Shown in Fig \ref{fig:orin_lidar_whisker_minmax1w1r}, SIM attains a mean of \textbf{0.2260}ms and tails p95=\textbf{0.2981}ms, p99=\textbf{0.3822}ms, max=\textbf{0.8788}ms. Relative to the best baseline using FAST DDS, p95 decreases by \textbf{70.02}\% and p99 by \textbf{64.80}\%; mean speedup=\textbf{3.02}$\times$. These reductions follow from R1/R3: no ROS message conversion and constant-time handoff on the sensor-to-application path. Unlike ROS, SIM allows for an additional scheduling priority elevation on the kernel level that can reduce the latency to achieve a mean of \textbf{0.0737}ms and tails p95=\textbf{0.2428}ms, p99=\textbf{0.2556}ms, max=\textbf{0.2647}ms

\paragraph{1W$\rightarrow$10R.}

\begin{figure}[htbp]
    \centering
    \includegraphics[width=1.0\columnwidth]{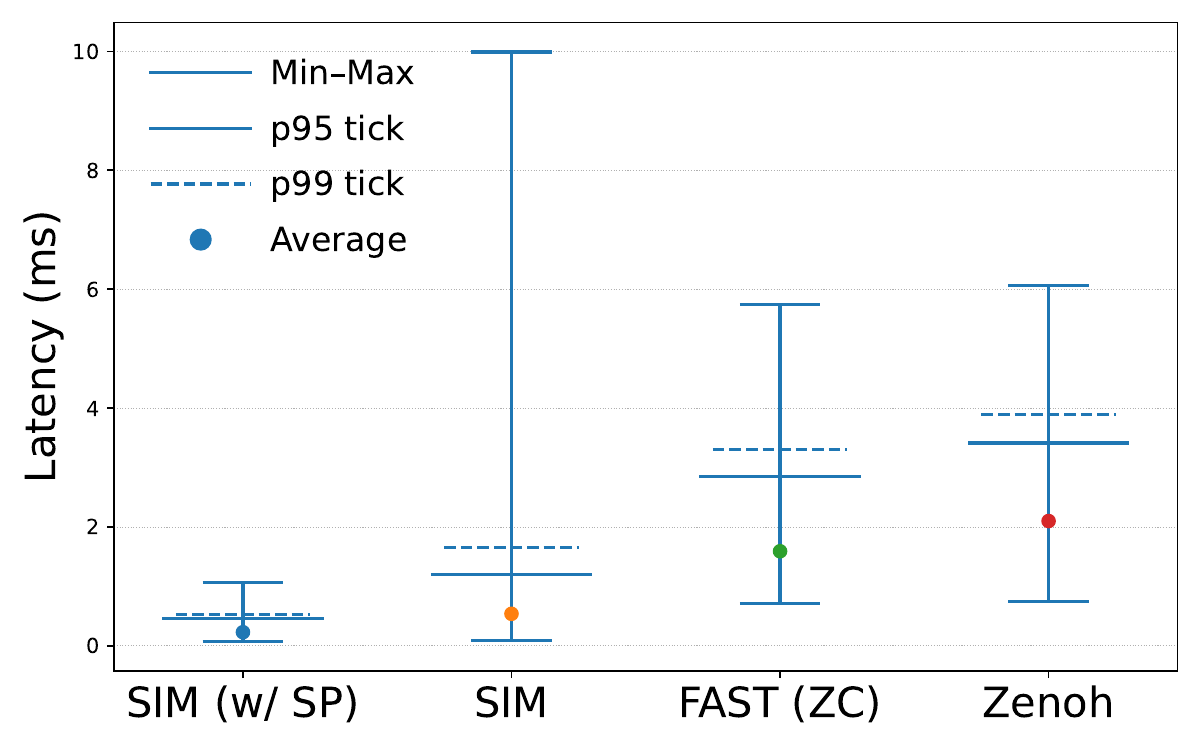}
    \caption{LiDAR latency (1W$\rightarrow$10R, linear-scale), Orin Nano. Whiskers show \emph{min–max}; filled dot is the \emph{mean}; long tick is \emph{p95}; short dashed tick is \emph{p99}. Transport-only; best-effort QoS; zero-copy enabled; pinned threads; fixed clocks; No scheduling priority elevation. \textbf{Mean (95\% CI across runs, $n=5$):} SIM 0.542 [0.513, 0.570]; Fast DDS 1.593 [1.583, 1.603]; Zenoh 2.136 [2.077, 2.196]. When scheduling priority is set to 99 with \emph{SCHED\_FIFO}, SIM's mean latency decreases to 0.231 [0.215, 0.246]. }
    \label{fig:orin_lidar_whisker_minmax1w10r}
\end{figure}

Shown in Fig \ref{fig:orin_lidar_whisker_minmax1w10r}, SIM attains a mean of \textbf{0.542}ms and tails p95=\textbf{0.513}ms, p99=\textbf{0.570}ms, max=\textbf{9.990}ms. Relative to the best baseline using FAST DDS, p95 decreases by \textbf{57.87}\% and p99 by \textbf{49.72}\%; mean speedup=\textbf{2.94}$\times$, supporting H2. Bounded overwrite-on-update (R2) avoids backlog under fan-out. While it seems like SIM has a larger maximum communication latency, unlike ROS, SIM allows for an additional scheduling priority elevation on the kernel level that can reduce the latency to achieve a mean of \textbf{0.0776}ms and tails p95=\textbf{0.2981}ms, p99=\textbf{0.3822}ms, max=\textbf{0.8788}ms

\subsubsection{Camera Results (640$\times$480 @ 30\,FPS)}
\paragraph{1W$\rightarrow$1R.}

\begin{figure}[htbp]
    \centering
    \includegraphics[width=1.0\columnwidth]{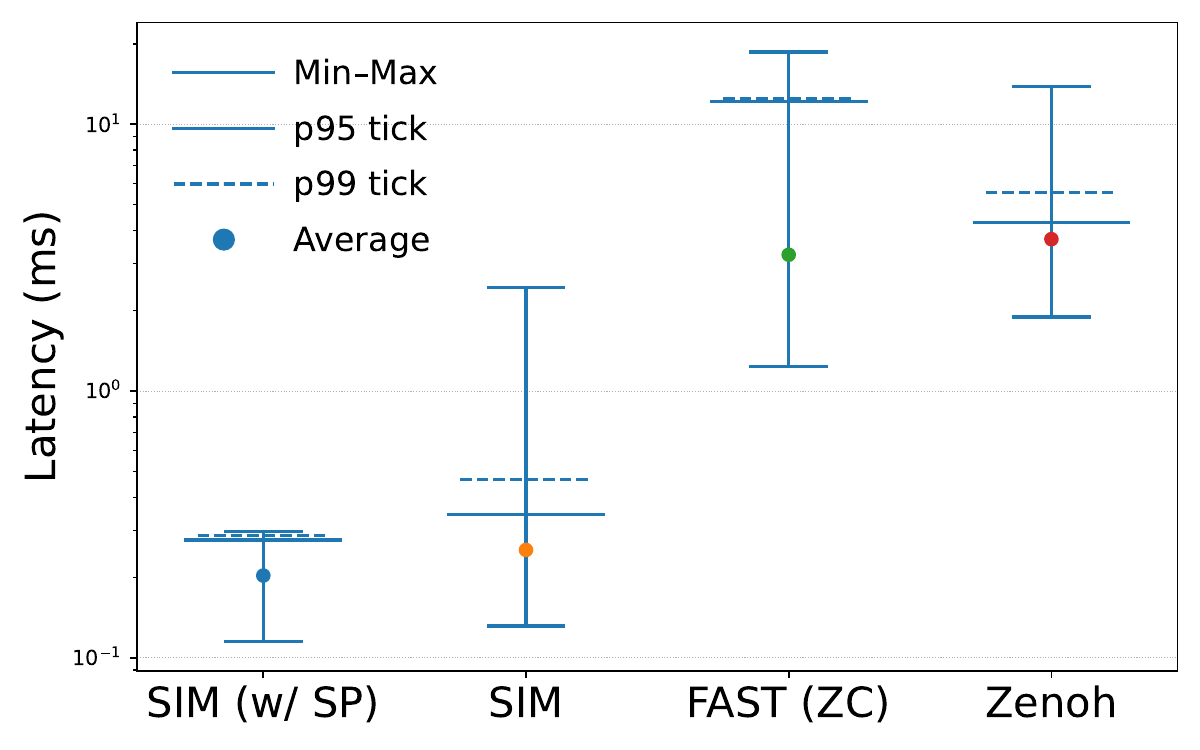}
    \caption{Camera latency (1W$\rightarrow$1R, log-scale), Orin Nano. Whiskers show \emph{min–max}; filled dot is the \emph{mean}; long tick is \emph{p95}; short dashed tick is \emph{p99}. Transport-only; best-effort QoS; zero-copy enabled; pinned threads; fixed clocks; No scheduling priority elevation. \textbf{Mean (95\% CI across runs, $n=5$):} SIM 0.253 [0.244, 0.261]; Fast DDS 3.40 [2.881, 3.920]; Zenoh 3.984 [3.877, 4.091]. When scheduling priority is set to 99 with \emph{SCHED\_FIFO}, SIM's mean latency decreases to 0.204 [0.195, 0.212]. }
    \label{fig:orin_camera_whisker_minmax1w1r}
\end{figure}

Shown in Fig \ref{fig:orin_camera_whisker_minmax1w1r}, SIM attains a mean of \textbf{0.2537}ms and tails p95=\textbf{0.3454}ms, p99=\textbf{0.4665}ms, max=\textbf{2.4371}ms. Relative to the best baseline using Zenoh, p95 decreases by \textbf{91.92}\% and p99 by \textbf{91.61}\%; mean speedup=\textbf{14.6}$\times$. These reductions follow from R1/R3: no ROS message conversion and constant-time handoff on the sensor-to-application path. Unlike ROS, SIM allows for an additional scheduling priority elevation on the kernel level that can reduce the latency to achieve a mean of \textbf{0.1152}ms and tails p95=\textbf{0.2765}ms, p99=\textbf{0.2872}ms, max=\textbf{0.2985}ms. When using SIM or Zenoh, all of the published messages were received, FAST DDS loses 2.0\% of the messages on average.

\paragraph{1W$\rightarrow$10R.}

\begin{figure}[!htbp]
    \centering
    \includegraphics[width=1.0\columnwidth]{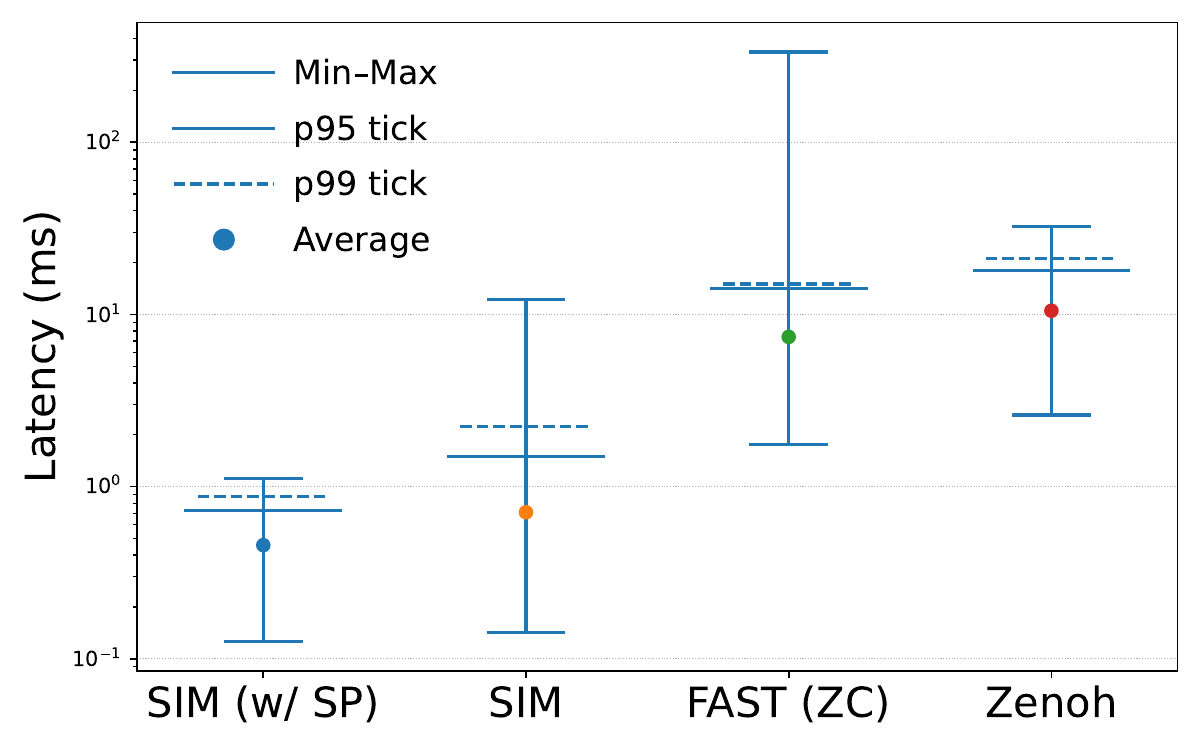}
    \caption{Camera latency (1W$\rightarrow$10R, log-scale), Orin Nano. Whiskers show \emph{min–max}; filled dot is the \emph{mean}; long tick is \emph{p95}; short dashed tick is \emph{p99}. Transport-only; best-effort QoS; zero-copy enabled; pinned threads; fixed clocks; No scheduling priority elevation. \textbf{Mean (95\% CI across runs, $n=5$):} SIM 0.717 [0.696, 0.739]; Fast DDS 7.226 [6.770, 7.682]; Zenoh 10.484 [10.176, 10.792]. When scheduling priority is set to 99 with \emph{SCHED\_FIFO}, SIM's mean latency decreases to 0.457 [0.126, 0.797].}
    \label{fig:orin_camera_whisker_minmax1w10r}
\end{figure}

Shown in Fig \ref{fig:orin_camera_whisker_minmax1w10r}, SIM attains a mean of \textbf{0.542}ms and tails p95=\textbf{0.513}ms, p99=\textbf{0.570}ms, max=\textbf{9.990}ms. Relative to the best baseline using FAST DDS, p95 decreases by \textbf{57.87}\% and p99 by \textbf{49.72}\%; mean speedup=\textbf{2.94}$\times$, supporting H2. Bounded overwrite-on-update (R2) avoids backlog under fan-out. While it seems like SIM has a larger maximum communication latency, unlike ROS, SIM allows for an additional scheduling priority elevation on the kernel level that can reduce the latency to achieve a mean of \textbf{0.4566}ms and tails p95=\textbf{0.7296}ms, p99=\textbf{0.8776}ms, max=\textbf{1.1149}ms. When using SIM or Zenoh, all of the published messages were received, FAST DDS loses 2.4\% of the messages on average.

\subsubsection{System Scheduler Load}
We report \emph{system-wide} scheduler load percentages (idle baseline, 1W$\rightarrow$1R, and 1W$\rightarrow$10R) measured over \textbf{1000} sensor data frames at \textbf{100}ms intervals, using identical affinities and fixed clocks across transports. The results are shown in Table \ref{tab:orin_sched}. Higher values indicate more time spent in kernel scheduling activity on the cores involved in the experiment.
\begin{table}[!ht]
\centering
\caption{System-wide scheduler load (\%). Same setup as latency tables; measured at rest and under LiDAR/camera workloads.}
\label{tab:orin_sched}
\begin{tabular}{l|c|c|c}
\toprule
\textbf{Condition} & \textbf{SIM} \% & \textbf{Fast DDS (ZC)} \% & \textbf{Zenoh} \% \\
\midrule
System Resting (idle)            & 6  & 6  & 6 \\
LiDAR 1W$\rightarrow$1R          & 32     & 123     & \textbf{25} \\
LiDAR 1W$\rightarrow$10R         & 139    & 60    & \textbf{58} \\
Camera 1W$\rightarrow$1R         & 68     & \textbf{40}     & 56 \\
Camera 1W$\rightarrow$10R        & 190    & \textbf{111}    & 114 \\
\bottomrule
\end{tabular}
\end{table}

We also report the average CPU usage measured over 1000 frames at a sampling interval of 100ms, using identical affinities and fixed clocks across transports. Higher values indicate higher usage of CPU resources involved in the experiment.The results are shown in Table \ref{tab:orin_cpu}

\begin{table}[!ht]
\centering
\caption{System-wide CPU usage (\%). Same setup as latency tables; measured at rest and under LiDAR/camera workloads.}
\label{tab:orin_cpu}
\begin{tabular}{l|c|c|c}
\toprule
\textbf{Condition} & \textbf{SIM} \% & \textbf{Fast DDS (ZC)} \% & \textbf{Zenoh} \% \\
\midrule
LiDAR 1W$\rightarrow$1R          & \textbf{5.2}     & 6.8     & 6.7 \\
LiDAR 1W$\rightarrow$10R         & 6.9    & 7.0    & \textbf{6.6} \\
Camera 1W$\rightarrow$1R         & 5.5     & 5.4     & \textbf{5.3} \\
Camera 1W$\rightarrow$10R        & 5.7    & 5.6    & \textbf{5.4} \\
\bottomrule
\end{tabular}
\end{table}

\noindent\textit{Interpretation.} SIM does not introduce significant overhead to the read/write process compared to other ROS~2 DDS, while also lowering mean and p95/p99. At equal QoS and concurrency, SIM achieves lower latency at slightly higher kernel scheduling cost and similiar CPU costs, aligning with R1--R3 from \S\ref{sec:design}. The increase in scheduler load and CPU usage is the direct result of the reader query the status of the shared memory region.

\noindent\textit{Fault injection.} Similar to FAST and Zenoh, terminating and resuming the writer mid-task does not have any effect on the readers due to the double buffer setup; terminating and resuming a reader mid-task does not have any effect on other readers or the writer. Corrupting the shared memory region will terminate both the writer and the reader process. The OS will assign a new shared memory region and resume normal read/write functions.

\subsection{Limitations}
Reported latencies measure \emph{transport only}; end-to-end perception-to-decision latencies include sensor capture and algorithm compute. All runs use best-effort QoS and enable zero-copy where available; switching to reliable QoS introduces retransmissions/back-pressure and increases baselines. Clock discipline and kernel build on Orin affect absolute numbers; we fix clocks, pin threads, and equalize affinity across stacks. Sensor formats and rates follow Table~\ref{tab: setup}. The comparative ordering follows from \S\ref{sec:design} (native layouts, no (de)serialization, bounded overwrite-on-update, constant-time publication) and should hold for equivalent intra-host settings.

\paragraph{Summary.}
Across LiDAR and camera workloads on Orin Nano, and under both 1W$\rightarrow$1R and 1W$\rightarrow$10R, SIM lowers mean and p95/p99 relative to Fast DDS (Zero-Copy) and Zenoh and maintains system-wide scheduler load. These results validate H1--H3 and directly reflect the mechanisms in \S\ref{sec:design}.

\subsection{Results on a Full-Scale Autonomous Vehicle}
\label{sec:vehicle}

\paragraph{Goal.}
We evaluate SIM \emph{in situ} on a production-ready autonomous vehicle running Autoware.Universe to answer three questions: (i) what is the end-to-end perception$\rightarrow$decision latency and CPU load under the stock ROS~2 pipeline; (ii) how much of that latency is attributable to \emph{communication} within ROS~2; and (iii) what changes when we replace only the \emph{localization} module’s intra-host transport with SIM, and what savings are achievable if SIM is applied across all intra-host stages.

\textbf{Transports and workloads.} We compare the sensor-to-decision latency of Autoware.Universe under native ROS 2~DDS (FAST (ZC)) to that of using \emph{SIM without elevate scheduling priority} to . Streams on autonomous vehicle (Table~\ref{tab: setup}): one outdoor LiDAR (64 channels @ 10\,Hz; \texttt{PointXYZ[]} with \textbf{115,200} points/frame) and seven camera (1920$\times$1200 @ 42-120\,FPS; \texttt{cv::Mat}).

\paragraph{Method.}
We instrument Autoware.Universe nodes with timestamping at module boundaries and collect transport timestamps at publisher commit and subscriber delivery (sensor capture and downstream actuation are excluded). For each run we record \textit{min}/\textit{mean}/\textit{p95}/\textit{p99}/\textit{max}/\textit{std} over \textbf{5} runs $\times$ \textbf{[10 minute runs}. CPU load is sampled as \emph{system CPU\%} and \emph{Autoware CPU\%} at \textbf{100}ms with fixed clocks and pinned threads. QoS is best-effort (depth=1) across all modules.

\paragraph{Baseline end-to-end and ROS~2 communication cost.}
Table~\ref{tab:veh_e2e} reports the end-to-end (E2E) perception$\rightarrow$decision latency under native Autoware.Universe. Table~\ref{tab:veh_comm} isolates the \emph{ROS~2 communication} latency measured across module boundaries (publisher commit to subscriber delivery), i.e., the portion SIM targets intra-host.

\begin{table}[!ht]
\centering
\caption{Vehicle: Autoware.Universe E2E perception$\rightarrow$decision latency (ms). Transport+compute on the vehicle; actuation excluded.}
\label{tab:veh_e2e}
\resizebox{0.5\textwidth}{!}{
\begin{tabular}{l|c|c|c|c|c}
\toprule
Metric & Total & Preprocessing & Localization & Perception & Planning\\
\midrule
Min  & 320.363 & 43.818 & 18.769 & 188.750 & 69.026 \\
Mean & 521.905 & 90.844 & 30.960 & 280.860 & 119.242\\
p95  & 788.794 & 127.421 & 66.822 & 396.179 & 198.372 \\
p99  & 1036.798 & 157.957 & 91.011 & 521.262 & 266.568 \\
Max  & 1417.972 & 250.467 & 260.475 & 514.595 & 392.435 \\
Std  & 123.079 & 25.847 & 17.822 & 38.677 & 40.734 \\
\bottomrule
\end{tabular}
}
\end{table}

\begin{table}[!ht]
\centering
\caption{Vehicle: ROS~2 intra-host communication latency (ms) across the Autoware.Universe pipeline.}
\label{tab:veh_comm}
\resizebox{0.5\textwidth}{!}{
\begin{tabular}{l|c|c|c|c|c}
\toprule
Metric & Total & Preprocessing & Localization & Perception & Planning\\
\midrule
Min  & 266.278 & 32.278 & 11.012 & 158.243 & 65.235 \\
Mean & 369.456 & 75.921 & 12.134 & 168.907 & 112.494\\
p95  & 522.480 & 102.380 & 27.526 & 207.145 & 184.800 \\
p99  & 692.367 & 118.247 & 43.785 & 208.300 & 248.32 \\
Max  & 917.147 & 170.147 & 199.210 & 282.8 & 341.69 \\
Std  & 66.511 & 19.825 & 17.822 & 1.401 & 37.085 \\
\bottomrule
\end{tabular}
}
\end{table}

\begin{figure}[htbp]
    \centering
    \includegraphics[width=1.0\columnwidth]{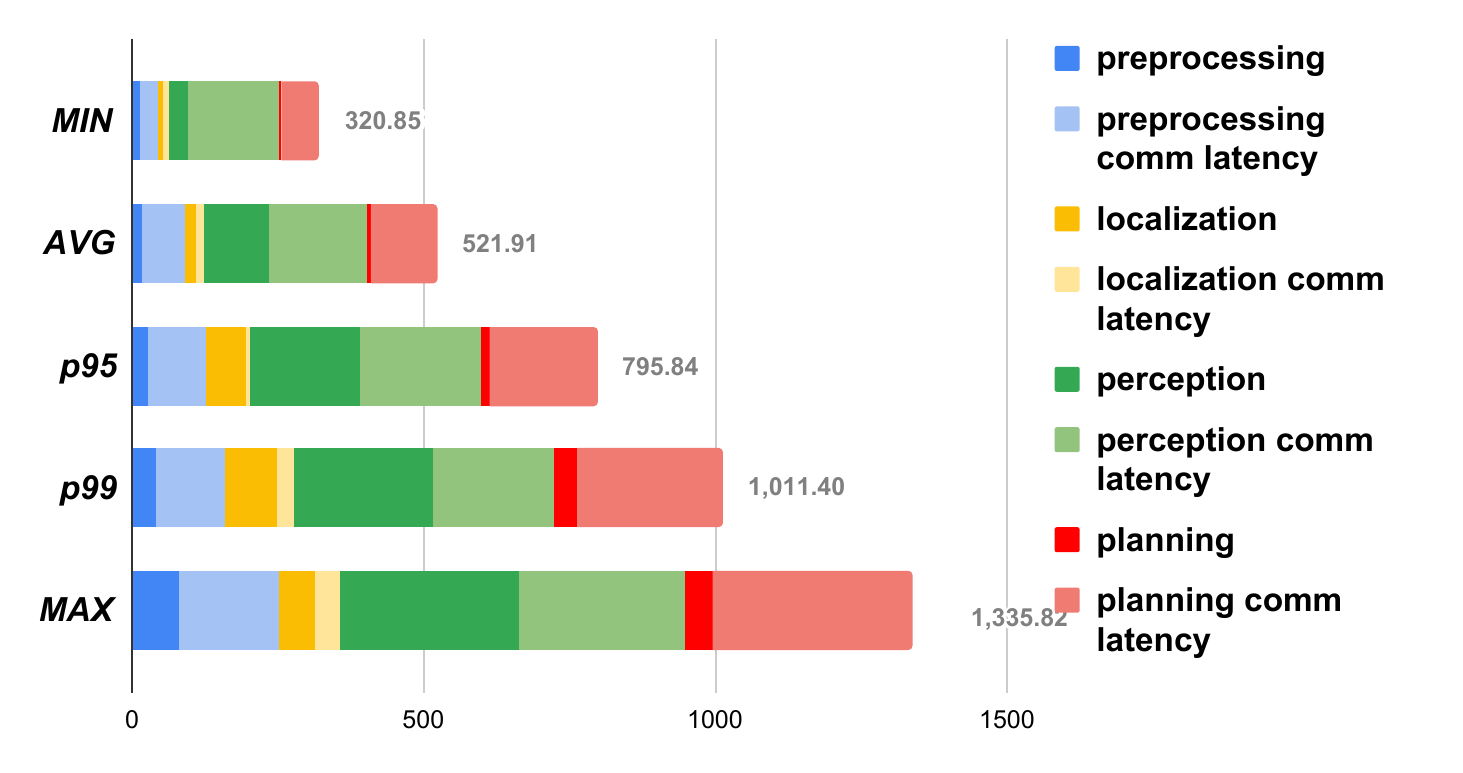}
    \caption{End-to-End latency breakdown on production-ready L4 autonomous vehicle running Autoware.Universe(March 2024 release). In preprocessing and planning tasks, the communication latency transferring the data to downstream critical applications is much larger than the task itself. In perception tasks, the communication latency takes about the same amount of time used by the detection-tracking algorithms.}
    \label{fig: e2ebreakdown}
\end{figure}

\paragraph{CPU load (system and Autoware).}
Table~\ref{tab:veh_cpu} summarizes CPU load with the stock stack and after replacing localization’s intra-host transport with SIM.

\begin{table}[!ht]
\centering
\caption{Vehicle: System and CPU load (\%). System-wide under fixed clocks and pinned threads.}
\label{tab:veh_cpu}
\begin{tabular}{l|c|c}
\toprule
Condition & System Load\% & CPU Usage\% \\
\midrule
System resting (idle)                    & 4.2 & 200.4 \\
Stock ROS~2 (Autoware)          & 57.1 & 1474.1 \\
SIM ROS~2 (Autoware)           & 55.2 & 1474.4 \\
\bottomrule
\end{tabular}
\end{table}

\paragraph{Replacing localization with SIM.}
We replace the \emph{localization} module’s intra-host transport with SIM. The NDT output frequency increases from \textbf{7.5\,Hz} to \textbf{9.5\,Hz} (the module’s configured maximum is 10\,Hz). Table~\ref{tab:veh_comm} reports localization-path transport latency before/after. As shown in Table \ref{tab:veh_comm}, when using ROS~2 FAST DDS (ZC), the total latency averages 121.804\,ms (8.2\,Hz), and the sensor publishes data at  9.5-10\,Hz. Over 10-minute runs, the average localization output drops to 7.5\.Hz. When using SIM to transport the data, the total latency decreases to sub-100\,ms and the localization module is able to keep up with the sensor output to prevent any backlog. 

\begin{table}[!ht]
\centering
\caption{Vehicle: E2E perception$\rightarrow$decision latency (ms) saved with SIM.}
\label{tab:veh_e2e_sim}
\resizebox{0.5\textwidth}{!}{
\begin{tabular}{l|c|c|c|c|c|c}
\toprule
 & Min & Mean & p95 & p99 & Max & Std \\
\midrule
\shortstack{Time Saved\\with SIM (ms)} & 218.750 & 231.649 & 219.855 & 307.594 & 318.031 & 53.681 \\
\bottomrule
\end{tabular}
}
\end{table}

\paragraph{End-to-end impact with SIM across intra-host stages.}
Applying SIM across all intra-host communications in \emph{preprocessing, perception, localization, and planning} yields an average E2E reduction of \textbf{231.65\,ms} (measured/estimated from per-link communication costs in Table~\ref{tab:veh_comm} and per-module paths). This corresponds to a shorter reaction distance of \textbf{13.6\,ft (4.1\,m)} at \textbf{40\,mph (64\,kmh)} and a shorter reaction distance of \textbf{23.8\,ft (7.3\,m)} at \textbf{70\,mph (112\,kmh)} on dry concrete\footnote{Reported as measured/estimated for this platform. Reaction-distance scaling uses $d = v \cdot \Delta t$ with speed in m/s and $\Delta t$ in seconds.} This reduction enlarges the safety margin available to the controller under emergency braking and high-speed maneuvers.

\paragraph{Limitations (vehicle).}
We report single-run distributions without confidence intervals; results reflect the tested vehicle, sensors, and kernel build. E2E latency includes module compute and intra-host communication but excludes actuation. QoS is best-effort and identical across configurations. Projections for “SIM across intra-host” are derived from measured per-link communication costs, concurrency is accounted for.

%% file: sections/5_Conclusion.tex
\section{Conclusion and Future Works} \label{sec:conclusion}



We introduced SIM, a domain-specific shared-memory framework for AVs that uses a lightweight, lock-free, double-buffered design to cut latency and jitter beyond general-purpose middleware. Evaluations on a Jetson Orin Nano and a full-scale AV show consistent end-to-end latency reductions for high-throughput LiDAR/camera workloads and up to ten readers, with gains up to an order of magnitude in some settings—improving perception-to-control responsiveness at higher speeds. We will release the SIM library and all benchmarking artifacts on GitHub upon acceptance.